\documentclass{article}

\usepackage{arxiv}

\usepackage[utf8]{inputenc} 
\usepackage[T1]{fontenc}    
\usepackage{hyperref}       
\usepackage{url}            
\usepackage{booktabs}       
\usepackage{amsfonts}       
\usepackage{nicefrac}       
\usepackage{microtype}      
\usepackage{lipsum}
\usepackage{graphicx}
\graphicspath{ {./images/} }

\usepackage{amsmath} 
\usepackage{amssymb}  
\usepackage{booktabs}
\usepackage{siunitx}
\usepackage{multirow}
\usepackage[dvipsnames]{xcolor}
\usepackage{hyperref}

\usepackage[export]{adjustbox}
\usepackage{algorithm}
\usepackage{algpseudocode}
\usepackage{marvosym}

\title{Lavender Autonomous Navigation with Semantic Segmentation at the Edge}

\author{Alessandro~Navone$^{1}$, Fabrizio~Romanelli$^{2}$, Marco~Ambrosio$ {1}$, Mauro~Martini${1}$, Simone Angarano$^{1}$ and Marcello Chiaberge$^{1}$ 
\thanks{$^{1}$ Department of Electronics and Telecommunications, Politecnico di Torino, 10129, Torino, Italy. \tt\footnotesize \{firstname.lastname\}@polito.it}}

\author{
 Alessandro Navone \\
  Department of Electronics and Telecommunications \\
  Politecnico di Torino\\
  Torino, TO, 10129 \\
  \texttt{alessandro.navone@polito.it} \\
   \And
 Fabrizio Romanelli \\
  Department of Civil and Computer Engineering \\
  University of Rome Tor Vergata\\
  Roma, RM, 00133 \\
  \texttt{fabrizio.romanelli@uniroma2.it} \\
   \And
Marco Ambrosio\\
  Department of Electronics and Telecommunications \\
  Politecnico di Torino\\
  Torino, TO, 10129 \\
  \texttt{marco.ambrosio@polito.it} \\
   \And
 Mauro Martini \\
  Department of Electronics and Telecommunications \\
  Politecnico di Torino\\
  Torino, TO, 10129 \\
  \texttt{mauro.martini@polito.it} \\
  \And
 Simone Angarano \\
  Department of Electronics and Telecommunications \\
  Politecnico di Torino\\
  Torino, TO, 10129 \\
  \texttt{simone.angarano@polito.it} \\
  \And
 Marcello Chiaberge \\
  Department of Electronics and Telecommunications \\
  Politecnico di Torino\\
  Torino, TO, 10129 \\
  \texttt{marcello.chiaberge@polito.it} \\
}

\begin{document}
\maketitle
\begin{abstract}
Achieving success in agricultural activities heavily relies on precise navigation in row crop fields. Recently, segmentation-based navigation has emerged as a reliable technique when GPS-based localization is unavailable or higher accuracy is needed due to vegetation or unfavorable weather conditions. It also comes in handy when plants are growing rapidly and require an online adaptation of the navigation algorithm. This work applies a segmentation-based visual agnostic navigation algorithm to lavender fields, considering both simulation and real-world scenarios.
The effectiveness of this approach is validated through a wide set of experimental tests, which show the capability of the proposed solution to generalize over different scenarios and provide highly-reliable results.
\end{abstract}

\keywords{Autonomous Navigation \and Semantic Segmentation \and Precision Agriculture}

\section{Introduction}
In recent times, the food and farming industries have been facing a significant rise in global demand for food production, resulting in an increased need for resources \cite{calicioglu_future_2019}.  As a result, there is a growing requirement for new methods and technologies to enhance efficiency and productivity while also ensuring sustainability throughout the process.
As a consequence, farming industries focused their resources on developing new techniques to boost productivity and lower production costs, thereby reducing the need for labor-intensive human work \cite{zhai_decision_2020}.

The use of Deep Learning (DL) plays a significant role in the shift towards autonomous agents carrying out tasks in Agriculture 3.0 and 4.0 scenarios. Its ability to analyze data from various sources and reduce the likelihood of human errors makes it a valuable tool that can generalize and reduce tedious work \cite{mavridou2019}.
DL has been proven to be beneficial for autonomous guidance systems, which include automating tasks such as navigation along the fields \cite{cerrato2021deep}, including localization \cite{winterhalter2021localization}, global path planning \cite{salvetti2023waypoint} and local motion planning \cite{martini2022}, \cite{navone2023autonomous}. Moreover, among the different kinds of farming techniques, it emerged how row-organized crops are the most widely adopted farming arrangement, representing around 75\% of the USA farmland \cite{Bigelow}.

The aim of this research is to address issues that arise when traditional localization methods, like GPS, are inaccurate due to factors such as weather conditions, obstructions, or signal interference caused by tall vegetation.
Robust navigation in fields can be considered the starting point to carry out more complex and structural tasks such as crop monitoring \cite{comba2019monitor}, diseases detection \cite{ferentinos_deep_2018} \cite{shruthi2019diseases}, spraying pesticides in a more localized and efficient way \cite{deshmukh2021}, fruit counting \cite{mazzia2020}, harvesting \cite{Droukas_2023} and monitoring the crop status and eventual diseases \cite{ferentinos_deep_2018} \cite{shruthi2019diseases}.
Within this context, autonomous guidance systems and artificial intelligence have a primary role in achieving the objectives above.

The aim of this project is to create a motion planner that can navigate through plant rows of medium to high height. Many existing examples face challenges such as high costs and limited scalability. Typically, autonomous systems that navigate through row crops use high-precision GPS receivers and accuracy enhancement techniques \cite{olivier2015}, or a combination of sensors including lasers and GPS \cite{moorehead2012}. However, vegetation can interfere with the GPS signal reducing its accuracy and reliability \cite{kabir2016}. 
Recent solutions involve the employment of multiple sensors such as GPS, inertial navigation systems (INS) wheels' encoder and LIDARs to improve localization accuracy \cite{astolfi2018}; however, equipment composed of multiple sensors can lead to an increase in the costs of the platform. A visual odometry system that uses a downward-looking camera to reduce costs was proposed in the study cited as \cite{zaman2019}. Nonetheless, the authors noted that accuracy tends to decrease when the path is longer due to the accumulation of odometric error, and therefore, to keep the error bounded, integration with an absolute reference is necessary.

The main contribution of this work can be identified as the extensive experimentation carried out to validate the proposed method, which proves its generalization capabilities from a simulation to a real-world environment.

The rest of this work can be summarized as follows: Section \ref{sec:methodology} introduces the adopted solution for visual-based navigation in row crops, starting from the segmentation model to the adopted control algorithm. Section \ref{sec:tests-and-results} presents the simulation environment and the experimental setup and, later, reports the obtained results for both simulation and real-world tests.
 Finally, Section \ref{sec:conclusions} redraws and comments on the main points of this work.

\section{Methodology}\label{sec:methodology}

\begin{figure}
    \centering
    \includegraphics[width=\textwidth]{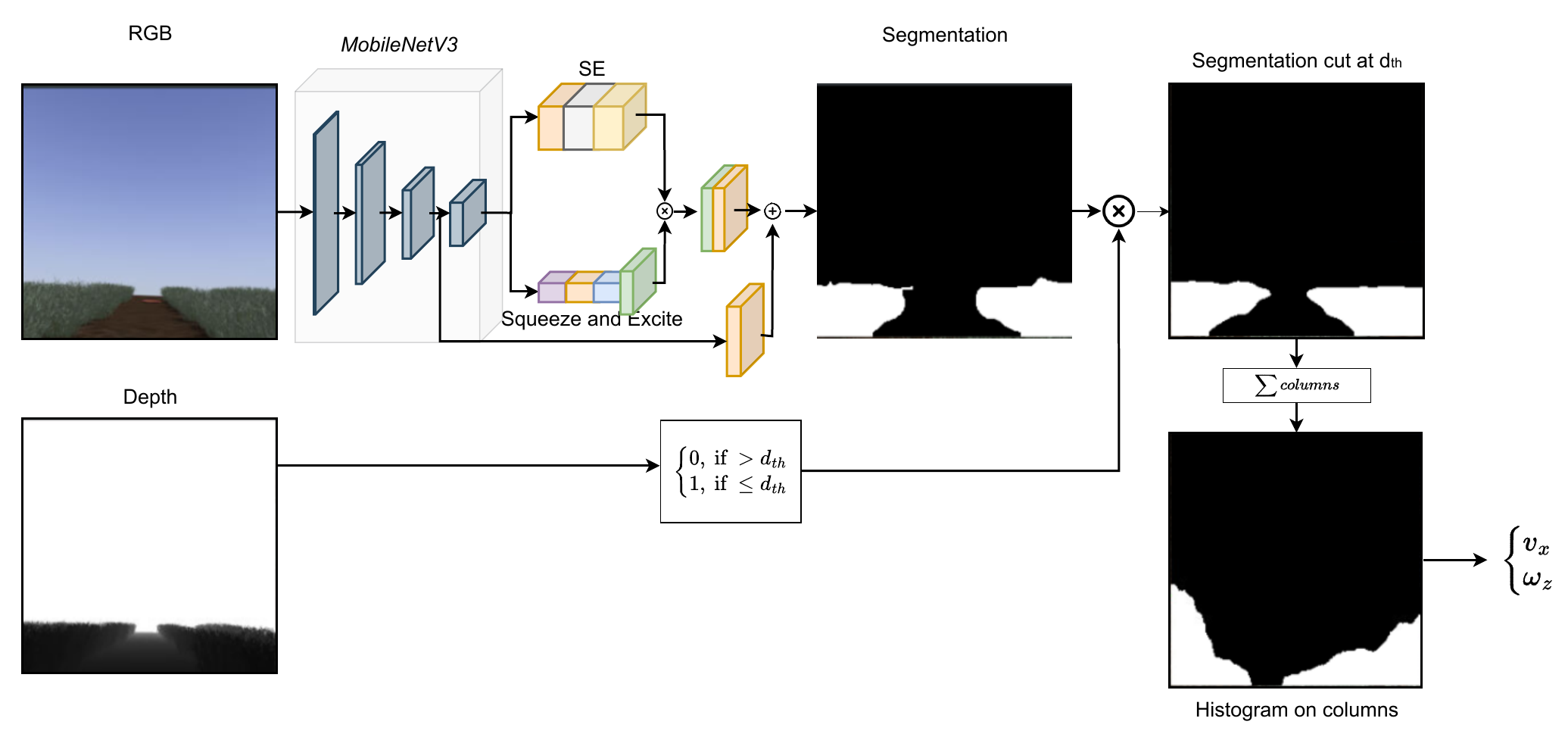}
    \label{fig:pipeline}
    \caption{Overall scheme of the pipeline of the proposed approach.}
\end{figure}

This study presents a new method for navigating without relying on position, using RGB-D data. The control algorithm created for this approach utilizes only real-time visual information to move through the rows of medium vegetation crops, like lavender. Additionally, this solution incorporates advancements in AI that enable "edge inferencing," where the robot's computer, often resource-limited, is able to process the input data in real time.
The basic idea behind this work consists of the generalization of the visual control algorithms proposed for high-vegetation crops in \cite{navone2023autonomous} to lower vegetation fields, obtaining a reliable and continuous control using a cost-effective set of sensors. In fact, the proposed system only exploits RGB and depth images, overcoming the problems experimented with using GPS-based solutions. Furthermore, the system can seamlessly integrate into a comprehensive navigation configuration, which includes a waypoints generator, a global path planner, and a classic navigation system based on a Dynamic Window Approach (DWA) that controls the robots outside of the rows \cite{cerrato2021deep}.

At instant $t$ camera located in the front of the UGV platform captures the RGB frame, denoted as $\mathbf{X}^{t}_{rgb} \in \mathbb{R}^{h \times w \times c}$, and the depth frame denoted as $\mathbf{X}^{t}_{depth} \in \mathbb{R}^{h \times w}$. Their dimensions are determined by the number of rows, represented by $h$, the number of columns, represented by $w$, and the number of channels, represented by $c$.
The RGB frame is later segmented semantically through a segmentation neural network, indicated as $H(\cdot)$. Its output, namely $\hat{\mathbf{X}}^{t}_{seg}$, is a binary segmentation mask that indicates the segmentation of the crop vegetation with ones, while the remaining areas of the frame are represented by zeroes.

\begin{equation}
    \label{eq:nn}
    \hat{\textbf{X}}^{t}_{seg} = H\left(\mathbf{X}^{t}_{rgb}\right)
\end{equation}

To achieve a more accurate determination of the end of the row's position, we eliminate noise from the segmented frame. We assess each column of the frame, denoted as $\hat{\textbf{X}}^{t}_{seg}(:,j)$, and if over 97 \% of the column is recognized as background, we set the remaining portion to zero.
Moreover, to further increase the algorithm's robustness, the segmentation masks of the last $N$, i.e. $\{t-N, \dots, t\}$ time instants, are super-imposed, obtaining a cumulative segmentation mask.

\begin{equation}
    \label{eq:mask_sum}
    \hat{\mathbf{X}}_{CumSeg}^{t} = \bigcup_{j=t-N}^{t} \hat{\textbf{X}}_{seg}^{t}
\end{equation}

where $ \hat{\mathbf{X}}_{CumSeg}^{t}$ is the cumulative segmentation mask, and $\bigcup$ represents the bitwise OR operation between the several segmentation masks.

Later on, the depth frame $\mathbf{X}_{d}^{t}$ is used to cut the cumulative segmented frame at a threshold distance $d_{th}$ in order to ignore the further segmentation data and better identify the continuation of the row.

\begin{equation}
    \label{eq:depth}
    \mathbf{\hat{X}}_{SegDepth}^{t}(i, j) = 
    \begin{cases}
    0, \textrm{  if  } \hat{\textbf{X}}^{t}_{CumDepth(i, j)} \cdot \textbf{X}_{d(i, j)}^{t}  > d_{th}\\
    1, \textrm{  if  } \hat{\textbf{X}}^{t}_{CumDepth(i, j)} \cdot \textbf{X}_{d(i, j)}^{t}  \leq d_{th}
    \end{cases}
\end{equation}

where $i = 0, \dots, h$, $j = 0, \dots, w$ and
$\mathbf{\hat{X}}_{SegDepth}^{t}$ is the segmentation frame cut with the depth information. 

In order to determine the center of the row and generate velocity commands, the columns of the depth-cut segmented image are summed, obtaining a histogram, $\textbf{h}^t$, obtained as in Equation \ref{eq:histo}. The practical idea behind this is to estimate the amount of vegetation per column.

\begin{equation}
    \label{eq:histo}
    \textbf{h}^{t}_{j} = \sum_{i = 1}^{w} \hat{\textbf{X}}_{SegDepth}^{t} (i, j) 
\end{equation}

Therefore, after obtaining the histogram $\textbf{h}_j^t$, it is evident how empty regions, namely clusters of zeros, represent regions in the field of view of the camera where no vegetation is present. Thus, identifying the widest cluster of zeros equals finding the continuation of the row.
Therefore, the desired cluster is identified with the following steps:
\begin{enumerate}
    \item The zeroes in the histogram are grouped in several clusters if they are in contiguous positions.
    \item If clusters are smaller than a threshold (i. e. smaller than 3 elements), they are discarded.
    \item the largest cluster is considered.
    \item If the largest cluster occupies more than 80\% of the space, it is considered an end-of-row condition.
\end{enumerate}
Once the desired cluster is identified, the distance $d$ between the center of the frame and the cluster center is considered and employed to calculate velocity commands.


\subsection{Segmentation Network}

Our real-time crop segmentation approach is based on previous works \cite{navone2023autonomous}, \cite{aghi2021}, that utilized a network consisting of a MobilenetV3 backbone for feature extraction and an efficient LR-ASPP segmentation head, as represented in Figure~\ref{fig:pipeline}. The LR-ASPP utilizes depth-wise convolutions, channel-wise attention, and residual skip connections to ensure a balance between accuracy and inference speed.

\begin{figure}[t]
    \setlength{\fboxsep}{0pt}
    \centering
    \begin{tabular}{c c c}
        & RGB & Segmentation Mask\\
        \vspace{2pt}
        (a) &\includegraphics[width=0.3\textwidth]{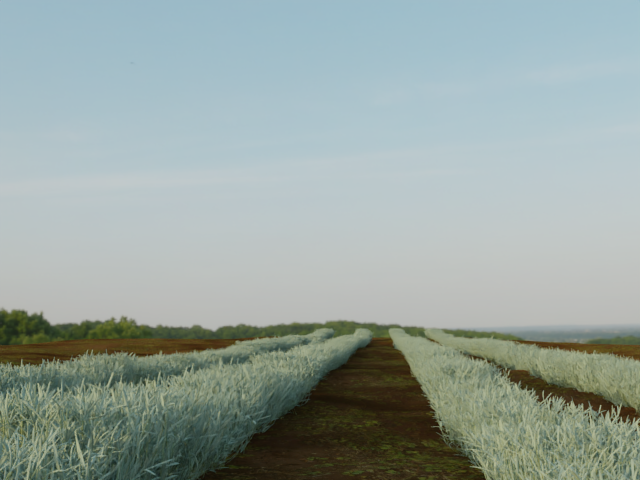} &
        \includegraphics[width=0.3\textwidth]{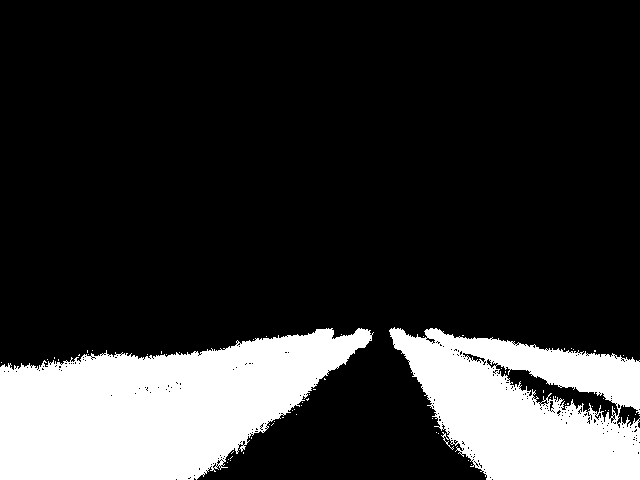}\\
        \vspace{2pt}
        (b) &\includegraphics[width=0.3\textwidth]{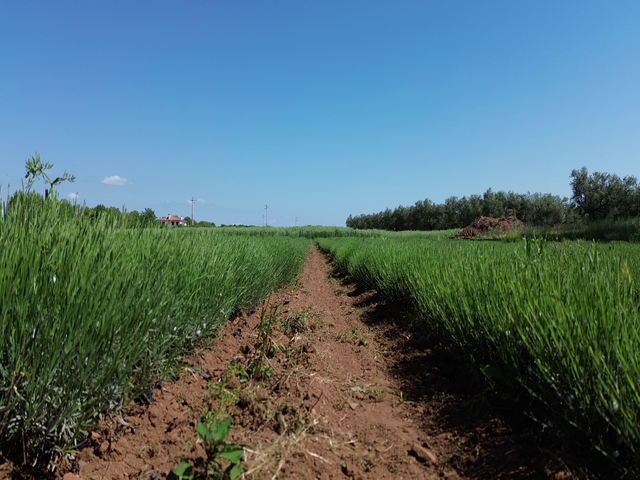} &
        \includegraphics[width=0.3\textwidth]{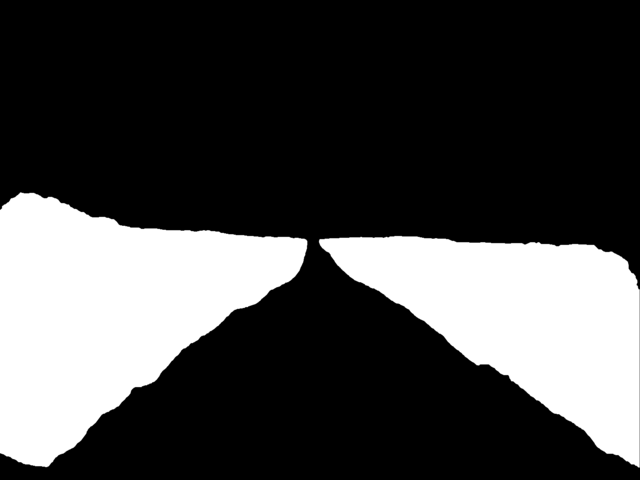}
    \end{tabular}
    \caption{Sample frames and relative segmentation masks from AgriSeg dataset including synthetic data (a) and real data (b).}
    \label{fig:dataset}
\end{figure}

\subsection{Velocity Command Generation}
Once the center of the largest cluster is identified and the distance $d$ from the center of the frame is evaluated, the velocity control commands can be generated. Furthermore, parabolic functions are used to generate the velocity functions.

\begin{equation}
    \label{eq:lin_vel}
    v_x = v_{x, max} \cdot \left(1 - \dfrac{d^2}{\left(\dfrac{w}{2}\right)^2}\right)
\end{equation}

\begin{equation}
    \label{eq:ang_vel}
    \omega_z = - \omega_{z, max} \cdot sign(d) \cdot \dfrac{d^2}{\left(\dfrac{w}{2}\right)^2}
\end{equation}

In \ref{eq:lin_vel} and \ref{eq:ang_vel}, $v_{x, max}$ and $\omega_{z, max}$ indicate the maximum linear and angular velocities respectively. The function $sign(\cdot)$ is used to determine the sign of a value, resulting in a value of 1 if the argument is greater than or equal to zero and -1 if it is less than zero.

\section{Tests and Results}
\label{sec:tests-and-results}
In this section, we present the results of experimental tests conducted both in simulation and in a real lavender field using a robot equipped with semantic segmentation capabilities at the edge. The objective of these tests was to evaluate the performance of the lavender autonomous navigation system and assess its effectiveness in identifying and avoiding obstacles. Specifically, the system has been tested on the field in order to assess its capability to be centered with respect to the rows of lavender as it travels along the row autonomously. 

\subsection{Evaluation Metrics}\label{sec:evaluation-metrics}

To quantitatively assess the performance of the lavender autonomous navigation system, we employed the following evaluation metrics:
\begin{itemize}
    \item Navigation Success Rate: This metric measures the success rate of the autonomous robot in navigating through the lavender field without colliding with any obstacles. It is calculated as the percentage of successful navigation trials out of the total number of trials conducted.
    \item Root Mean Square Error (RMSE) between the actual trajectory and the center of the rows: This metric measures the autonomous system's capability to maintain an equal distance from the rows while navigating toward the end of the field.
\end{itemize}
Regarding the navigation success rate metrics, we will report the percentage of successful navigation trials out of the total number of trials for the experiments in the next section. 

While for the simulation test, we have an absolute ground truth provided by the simulator, in the real-world experiments, we considered the effective ground truth of the robot path to be the data acquired from the GPS RTK sensor during the experimental runs. Furthermore, in order to get a more accurate ground truth, we resorted to a Visual Odometry system fusing the trajectory estimated by an optical flow algorithm and the one estimated by the ORB-SLAM2 algorithm (a modified version of the one presented in~\cite{visualSensorFusion}). The GPS RTK and Visual Odometry paths are finally fused and used as the trajectory ground truth for all the experiments. In order to assess how close the robot trajectory is from the center of the lavender rows, a set of measurements have been performed, measuring the GPS RTK coordinates, with $1$~m step along the path, at the middle of the rows.

\subsection{Segmentation Network Training}\label{sec:segmentationNetworkTraining}
Frames with dimensions $w = 224$ and $h = 224$ were considered as input of the Segmentation Neural network. Moreover, the number of channels $c$ is equal to 3 since they are RGB images. The model was trained on a combination of synthetic and real images from the lavender section of the AgriSeg dataset \cite{martini2023enhancing}. The dataset consisted of 4800 synthetic images and 1100 real ones. The model underwent 50 epochs of training with an ADAM optimizer, utilizing a learning rate of $3 \cdot 10^{-4}$. The data was augmented through cropping, flipping, grayscaling, and random jitters. The training of the model was conducted using TensorFlow 2 environment, starting from an ImageNet pretrained network on a single Nvidia RTX 3090 GPU.

\subsection{Simulation Tests} \label{sec:sim-tests}

\subsubsection{Simulation Setup} \label{sec:sim-setup}
The software used to perform the simulation is Gazebo \footnote{\url{https://classic.gazebosim.org/}} since it is one of the most supported and diffused simulators for robotics applications. Blender \footnote{\url{https://www.blender.org/download/lts/3-3/}} is used to create realistic models of plants and terrain that are exported are assembled in a Gazebo world using a procedural tool \cite{martini2023enhancing}. The robot model used in the simulation is the Husky UGV, the same that is used in real tests to reduce the gap between simulation and reality. Simulations are performed in rows of $8-10$~m length, and at least three runs are performed to assess the repeatability of the control algorithm.

\subsubsection{Simulation Results} \label{sec:sim-results}
\begin{figure}
\centering
\includegraphics[width=1.\linewidth]{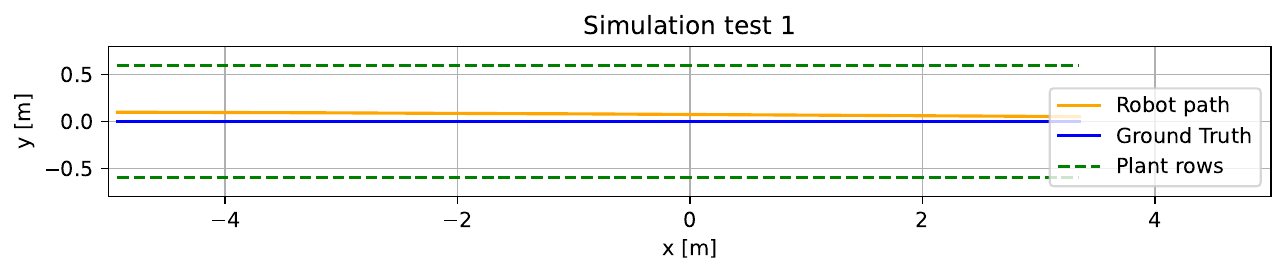}
\caption{Robot trajectories for one simulation run. The green dashed lines represent the plants, the blue line is the ideal central line, and the golden line represents the actual robot trajectory.}
\label{figure:simulation-res}
\end{figure}

The results obtained from the simulation are evaluated according to the metrics reported in Section~\ref{sec:evaluation-metrics}. The Navigation Success Rate is 1.0 since, in all the runs, the robot managed to reach the end of the row without hitting plants. The overall behavior of the robot was robust and reliable navigation until the end of the row. The RMSE of the robot trajectory with respect to the center for all the performed tests is reported in Table \ref{table:experimentalResults}. In addition, Figure \ref{figure:simulation-res} represent the trajectory followed by the robot during a simulation and the central line between the plants.

\subsection{Experimental Tests} \label{sec:experimental-tests}
\begin{figure}
\centering
\includegraphics[width=0.6\linewidth]{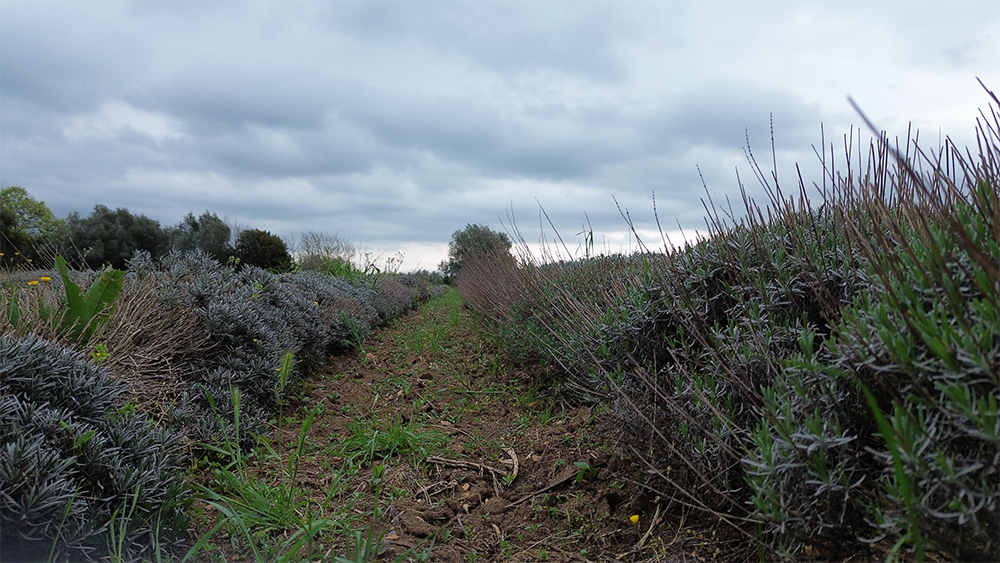}
\caption{A first-person view of the lavender rows with the path followed by the robot during the experimental tests.}
\label{fig:lightingConditions}
\end{figure}
\subsubsection{Experimental Setup}\label{sec:experimental-setup}

The experimental tests were conducted in a real lavender field with varying degrees of complexity in terms of terrain, vegetation density, and lighting conditions. The autonomous robot used in the experiments was equipped with state-of-the-art sensors, including cameras for capturing RGB images and depth information, and a powerful onboard processing unit capable of performing semantic segmentation at the edge.
The lavender field, where the experimental tests have been run, is located in Tuscania, Italy. In the same estate, two lavender fields are located. One field is composed of $22$ lavender rows with an average length of $35$~m with a distance between rows of about $80$~cm. The other field is composed of $16$ lavender rows with an average length of $25$~m with a distance between rows of about $65$~cm. The terrain was even, and there were weeds along the rows, even taller than the robot itself. The lighting conditions varied a lot between the experiments: from sunny to cloudy, also changing during every single run. An image of the environment is reported in Figure~\ref{fig:lightingConditions}, where a first-person view of the lavender rows is visible, together with the cloudy sky and weeds along the path.

\begin{figure}
\centering
\includegraphics[width=0.6\linewidth]{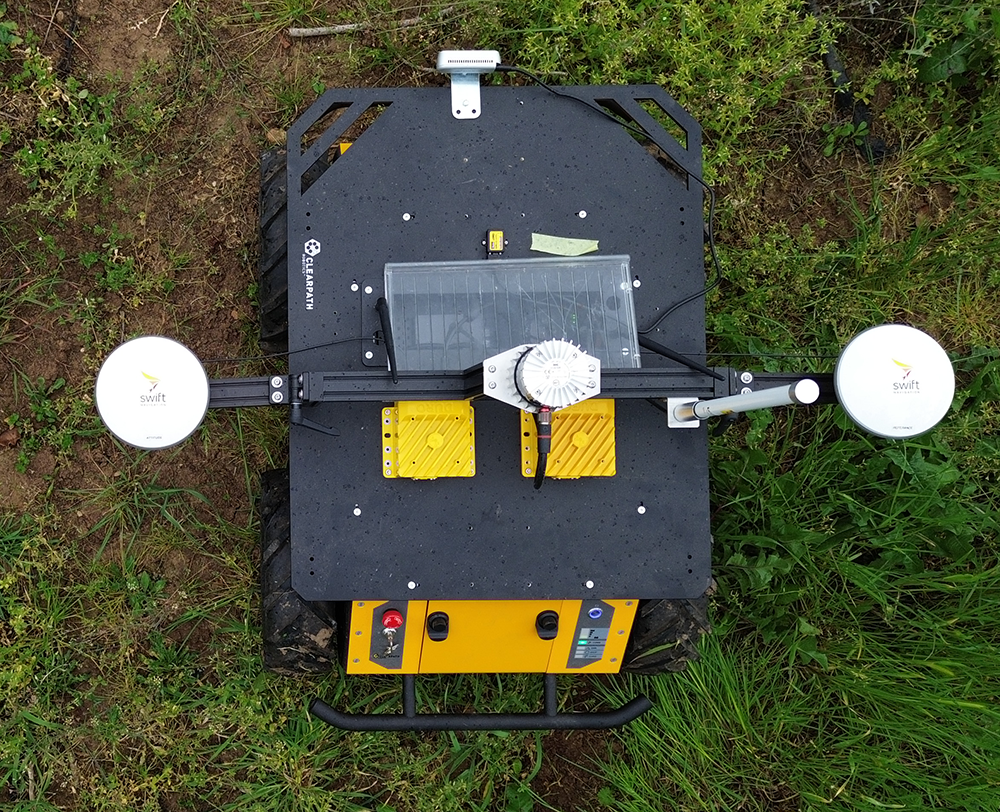}
\caption{An aerial view of the ClearPath Husky UGV, where the GPS antennas, IMU, LIDAR, and Intel RealSense D435 Camera are visible.}
\label{fig:robot}
\end{figure}

The robot used in the tests is a ClearPath Robotics Husky UGV, shown in Figure~\ref{fig:robot}. Husky is a medium-sized robotic development platform. It has a large payload capacity and power systems. Stereo cameras, LIDAR, GPS, and IMUs are mounted on the UGV in order to achieve a high level of autonomy. The Husky UGV has a rugged construction and high-torque drivetrain that enables the robot to work in harsh environments and uneven terrain. Husky is also fully supported in ROS, where the algorithms have been developed, tested, and used in real experiments. Its external dimensions are $990 \times 670 \times 390$~mm, while its internal dimensions are $296 \times 411 \times 155$~mm. It weighs $50$~Kg, and it can support a payload of $75$~Kg at $1$~m/s maximum speed with a $3$ hours battery autonomy.

The real experiments consist of the robot running autonomously between the lavender rows with the aim of maintaining a proper distance from the lavender plants exploiting the semantic segmentation algorithms presented in Section~\ref{sec:methodology}, while moving towards the end of the row.

\begin{table}
\centering
\caption{Results of simulation and real-world tests. In simulation tests, the RMSE is computed between the geometrical center of the row and the actual trajectory; in real-world tests, it is computed between the actual robot trajectory (from the fused VO and GPS trajectory) and the center of the row.}
\begin{tabular}{l l c c}
\toprule
Test number & Test type  & Path length {[}m{]} & RMSE {[}m{]}  \\ \midrule
test 1      & simulation & 8.26                & 0.077         \\
test 2       & simulation & 8.36                & 0.073         \\
test 3      & simulation & 8.14                & 0.082         \\ \midrule
overall     & simulation & 24.76               & 0.077 $\pm$ 0.004 \\ \midrule
test 1      & real world & 23.27               & 0.289         \\
test 2      & real world & 21.62               & 0.259         \\
test 3      & real world & 11.88               & 0.049         \\
test 4      & real world & 10.52             & 0.051         \\
test 5      & real world & 18.55               & 0.152         \\
test 6      & real world & 17.98               & 0.113         \\
test 7      & real world & 21.12               & 0.235         \\ \midrule
overall     & real world & 124.94               & 1.164 $\pm$ 0.098 \\ \bottomrule
\end{tabular}
\label{table:experimentalResults}

\end{table}

\subsubsection{Experimental Results}\label{sec:experimental-results}

During the experimental tests, the lavender autonomous navigation system exhibited robust performance and demonstrated promising results. Here, we present the key findings based on the evaluation metrics mentioned in Section~\ref{sec:evaluation-metrics}.

The autonomous robot achieved a remarkable navigation success rate of $70$\%. This indicates that the lavender autonomous navigation system effectively planned and executed collision-free paths, successfully maneuvering through the lavender field while avoiding obstacles. In order to assess this metric, we performed $10$ runs within different lavender rows with different lengths, and the robot successfully moved without any collision on $7$ tests. The $3$ runs where the robot failed to perform its task were characterized by strong changes in lighting conditions (e.g., the sky conditions changed during the experiment passing from cloudy to fully sunny). However, even in these cases, the robot was able to complete about $80$\% of the path.

Furthermore, in order to assess the system's ability to maintain a centered position with respect to the lavender rows, we measured the RMSE between the actual trajectory and the center of the rows. Here we report three significant results: one test for a long path (about $20$~m) that presented the highest RMSE, one for a long path (about $20$~m) that presented the lowest RMSE, and one for a short path of about $10$~m. These results are summarized in Table~\ref{table:experimentalResults}.
From these results, we noticed that the accuracy decreases when the path is longer due to the accumulation of odometric errors; also, tests 1 and 2 were performed along rows where the lavender plants were blooming, and therefore the rows were narrower, compared to those of test 3 were the lavender plants were lower. Finally, we present the details about the trajectories for the three tests just introduced.
\begin{figure}
\centering
\includegraphics[width=1.\linewidth]{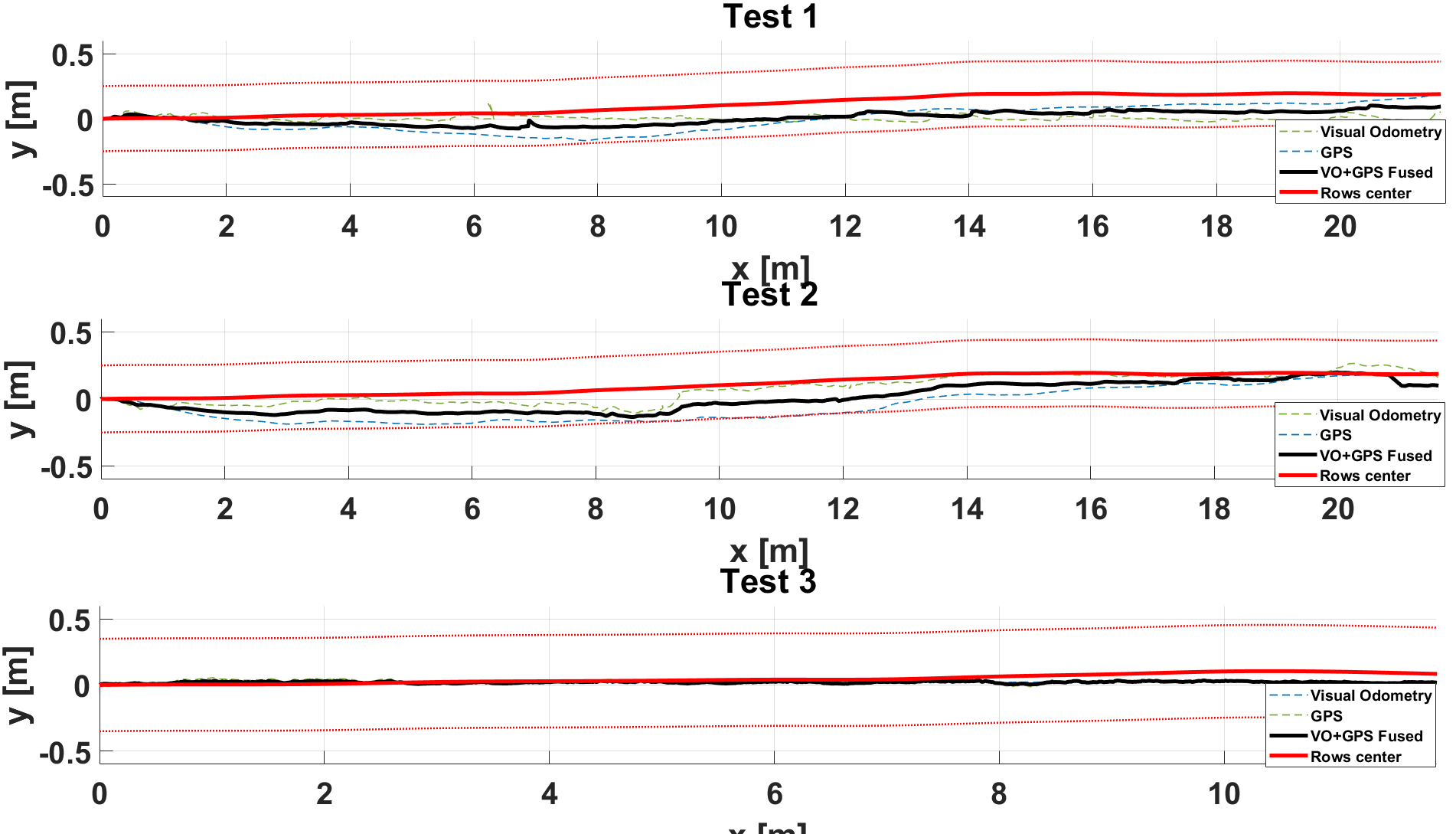}
\caption{Robot trajectories for three tests. The red solid line represents the center of the row, the green dashed line is the Visual Odometry trajectory, the blue dashed line is the GPS trajectory whilst the black dashed line represents the VO and GPS fused trajectory. For clarity's sake, we reported the boundaries of the lavender rows as red dotted lines.}
\label{fig:experimentalTests}
\end{figure}
Figure~\ref{fig:experimentalTests} shows the experimental results for the three tests; there, we reported, for each test, the Visual Odometry estimated trajectory, the GPS RTK trajectory, the VO and GPS fused trajectory, and the center of the row. In the same figure, we also reported the row boundaries as red dotted lines. These boundaries are taken with a rough estimation of the lavender plants' center and help to determine the surface that can be traveled between the rows. As previously pointed out, test number 3 shows the best results in terms of the ability to stay close to the row's center because of two factors: the path length and the lavender plant state (lower and smaller than in cases 1 and 2).

\section{Conclusions}
\label{sec:conclusions}

Overall, the experimental results demonstrate the efficacy of the lavender autonomous navigation system with semantic segmentation at the edge. The high accuracy in semantic segmentation showcases the system's potential for autonomous operation in real-world lavender farming environments. These results validate the effectiveness of the proposed system in enabling autonomous robots to navigate through lavender fields efficiently, reducing manual intervention, and enhancing productivity in lavender cultivation. However, further testing and refinement are necessary to ensure robustness and adaptability across a wider range of lavender field conditions and environmental variations.

\subsubsection{Acknowledgements} This work has been developed with the contribution of Politecnico di Torino Interdepartmental Center for Service Robotics PIC4SeR \footnote{\url{www.pic4ser.polito.it}}.

\bibliographystyle{unsrt}  
\bibliography{bibliography}  


\end{document}